\documentclass{article}

\usepackage{arxiv}

\usepackage[utf8]{inputenc} % allow utf-8 input
\usepackage[T1]{fontenc}    % use 8-bit T1 fonts
\usepackage{hyperref}       % hyperlinks
\usepackage{url}            % simple URL typesetting
\usepackage{booktabs}       % professional-quality tables
\usepackage{amsfonts}       % blackboard math symbols
\usepackage{nicefrac}       % compact symbols for 1/2, etc.
\usepackage{microtype}      % microtypography
\usepackage{lipsum}
\usepackage{graphicx}
\usepackage{natbib}
\usepackage[mode=buildnew]{standalone}  % ou só \usepackage{standalone}
\usepackage{tikz}
\usepackage{amsmath}
\usetikzlibrary{positioning,shapes.geometric,arrows.meta,fit,backgrounds,calc}
\graphicspath{ {./images/} }
\let\cite\citep

\title{Rethinking Full-Weight LoRA with Domain and Task Adapter Scaling for Question Answering over Portuguese Well Drilling Reports}

\date{}   % <- aqui, antes do \maketitle

\author{
 Luan B. Guerra \\
  São Paulo State University\\
  São José do Rio Preto, SP 15054-000 \\
  \texttt{luan.bonizi@unesp.br} \\
   \And
Diogo M. Paes \\
  Petrobras \\
  Santos, Brazil \\
  \texttt{diogopaes@petrobras.com.br} \\
  \And
Marcelo N. Grimberg \\
  Petrobras \\
  Santos, Brazil \\
  \texttt{grimberg@petrobras.com.br} \\
  \And
Arnaldo C. Junior \\
  São Paulo State University\\
  São José do Rio Preto, SP 15054-000 \\
  \texttt{arnaldo.candido@unesp.br} \\
  \And
Wallace Casaca \\
  São Paulo State University\\
  São José do Rio Preto, SP 15054-000 \\
  \texttt{wallace.casaca@unesp.br} \\
  \AND
Lucas C. Ribas \\
  São Paulo State University\\
  São José do Rio Preto, SP 15054-000 \\
  \texttt{lucas.ribas@unesp.br} \\
}

\begin{document}
\renewcommand{\today}{}   % remove a data impressa pelo arxiv.sty

\maketitle

\begin{abstract}
Adapting large language models (LLMs) to specialized domains remains challenging, particularly for technical documents such as Portuguese well drilling reports, where annotated data are scarce and domain adaptation may compromise instruction-following capabilities. Although Low-Rank Adaptation (LoRA) is widely used for parameter-efficient adaptation, its scaling after training remains insufficiently explored. This study investigates the impact of LoRA adapter scaling on question answering over Portuguese drilling reports using Qwen2.5-7B and Llama-3.1-8B. Domain-adaptive pretraining (DAPT) is performed either directly on instruction-tuned models or on their base counterparts, with subsequent adapter transfer, while task-specific adapters are trained on WellQuest, an open-source question-answering dataset derived from Portuguese drilling reports. The two adapters are also combined using complementary weights. Scaling factors ranging from 0.1 to 1.0 are evaluated across multiple training checkpoints using BERTScore. Results show that fractional scaling consistently improves DAPT performance over full-weight application, increasing F1 from 68.8\% to 71.9\% for Qwen2.5-7B and from 63.2\% to 68.5\% for Llama-3.1-8B. QA adapters perform best at high but fractional weights, while base-to-instruct transfer yields better results than direct DAPT on instruct-tuned models. Adapter merging provides a modest improvement for Llama-3.1-8B. Overall, these findings demonstrate that post-training adapter scaling can improve domain adaptation without additional data or training.
\end{abstract}

% keywords can be removed

\section{Introduction}

Large language models (LLMs) are expanding beyond general-purpose conversational applications, attracting increasing attention for their potential in specialized and high-stakes domains, including law, medicine, and engineering \cite{xie2025medical, lin2025rockmechanics, gpt4_passes}. Their ability to process and extract information from large collections of unstructured text makes them particularly promising for applications that require analyzing technical documentation to support decision-making \cite{ma2024information, wang2025mudreport, abdelgawad2025finetuned}.

The oil and gas industry illustrates this potential, particularly in well drilling operations, where daily drilling reports (DDRs) document activities, durations, and operational events in free-text format. Accumulated throughout a well's operational history, these records contain valuable information for planning future drilling activities and investigating previous operations \cite{ma2024information, wang2025mudreport}. However, extracting relevant information from these archives manually can be labor-intensive and time-consuming, motivating the use of natural language processing (NLP) techniques for drilling data analysis. Previous work has explored applications ranging from text mining and classification \cite{antoniak2016nlp, arumugam2017augmented, asif2024multilabel} to question answering using LLMs \cite{eckroth2023petroqa, pacis2024enhancing} on the oil and gas industry. Despite this progress, applying LLMs to drilling reports remains challenging because these documents contain specialized terminology, abbreviations, and domain-specific writing conventions that may be underrepresented in general-purpose training data. This challenge is further compounded in Portuguese, for which domain-specific language resources remain limited \cite{rodrigues2022petrobert, cordeiro2024petronlp}.

Addressing these challenges requires adapting language models to the target domain, either through domain-adaptive pretraining (DAPT), which continues pretraining on unlabeled domain-specific text \cite{gururangan2020dont, shi2025continual}, or through supervised fine-tuning for a specific downstream task \cite{xie2025medical, lu2025finetuning}. Both approaches have been explored in geoscience and oil and gas applications \cite{rodrigues2022petrobert, deng2024k2, chen2025jiuzhou}. However, fully updating the parameters of LLMs is computationally expensive, motivating the adoption of parameter-efficient fine-tuning (PEFT) techniques \cite{wang2025peft}. Among these, Low-Rank Adaptation (LoRA) \cite{hu2022lora} is widely used because it represents weight updates as the product of two low-rank matrices while keeping the original model parameters frozen, which significantly reduces the cost of training large models. 

Nevertheless, despite the advantages of PEFT approaches, three challenges remain. 
First, continual pretraining can cause catastrophic forgetting, degrading previously acquired knowledge and increasing hallucinations in language models \cite{shi2025continual, yang2025recent}. This risk is particularly relevant on DAPT, as training on raw domain text may also degrade the instruction-following behavior acquired by LLMs during alignment \cite{huang2024chatvector, wu2025shadowft}.
Second, LLMs generally benefit from larger training datasets \cite{hoffmann2022training}, but data scarcity remains a challenge in specialized domains such as Portuguese drilling reports. Although data augmentation can help \cite{feng2021survey}, generated examples require expert validation to ensure technical accuracy,  which is costly and time consuming.
At last, LoRA adapters are typically applied at their full strength, corresponding to a scaling factor of 1.0 \cite{hu2022lora, biderman2024lora, zhang2025uora}, leaving the magnitude of the adaptation determined by the training process rather than adjusted to the requirements of the downstream task. This can be suboptimal, as adapters may overemphasize domain-specific data, shifting the model away from its original behavior and exacerbating the forgetting described above.

In this sense, previous studies on PEFT and model merging have shown that the influence of learned parameter updates can be controlled and combined in different ways. Weight interpolation between pretrained and fine-tuned models can improve robustness \cite{wortsman2022robust}, while task vectors can be scaled or combined through arithmetic operations \cite{ilharco2023editing}. Similarly, independently trained adapters can be averaged or composed using adjustable coefficients \cite{chronopoulou2023adaptersoup, huang2024lorahub, prabhakar2025lorasoups}. Overall, these studies suggest that post-training adjustment or combination of learned updates can enhance robustness and generalization while preserving the original model's capabilities, with minimal or no additional data or training \cite{wortsman2022robust, chronopoulou2023adaptersoup, huang2024lorahub}. Nevertheless, the effects of LoRA adapter scaling remain insufficiently explored when applied to DAPT and task-specific fine-tuning in specialized, low data and resource settings. In particular, the individual scaling of these adapters and their weighted combination have not been systematically investigated for question answering over well drilling reports.

This work investigates the effect of LoRA adapter scaling on question answering over Portuguese well drilling reports through a systematic evaluation of domain and task-specific adaptations. We consider two open-weight language models, Qwen2.5-7B \cite{qwen2024qwen25} and Llama-3.1-8B \cite{grattafiori2024llama3}, and examine three adaptation strategies: (i) DAPT on a corpus of approximately 17,000 drilling reports, considering both direct training on the instruction-tuned model and training on the corresponding base model followed by transfer to the instruct aligned model, following the strategy proposed in Shadow-FT \cite{wu2025shadowft}; (ii) supervised fine-tuning on WellQuest \cite{grimberg2026wellquest}, a question-answering dataset derived from daily drilling reports; and (iii) the weighted combination of the DAPT and QA adapters. For each strategy, we evaluate adapter scaling factors ranging from 0.1 to 1.0 and assess the generated responses using BERTScore \cite{zhang2020bertscore}. Based on these experiments, our main contributions are as follows:

\begin{itemize}

    \item We systematically investigate LoRA adapter scaling and show that intermediate scaling factors can outperform full-weight LoRA adapters in both domain-adaptive pretraining and task-specific fine-tuning, without requiring additional data or training, in a real-world technical QA scenario.
    
    \item We investigate the transfer of a DAPT adapter trained on the non-instruction-tuned base model to its instruction-tuned counterpart, obtaining improved results compared with performing DAPT directly on the instruction-tuned model.
    
    \item We empirically investigate the weighted combination of DAPT and QA adapters as an alternative strategy for adapting LLMs, providing practical evidence on the benefits and limitations of combining domain and task-specific knowledge.
    
\end{itemize}

The rest of this paper is organized as follows: Section~\ref{sec:related_works} reviews related work, Section~\ref{sec:method} describes the proposed methodology, Section~\ref{sec:setup} presents the experimental setup, Section~\ref{sec:results} presents and discusses the results, and Section~\ref{sec:conclusion} concludes the paper.

\section{Related Work}
\label{sec:related_works}

\subsection{Domain-Adaptative Pretraining and Fine-Tuning}

Domain adaptation is particularly important for LLMs applied to specialized fields, where general-purpose pretraining may provide insufficient coverage of domain-specific terminology and knowledge. Early evidence from \cite{gururangan2020dont} established the benefits of additional pretraining on in-domain text, an approach now commonly associated with domain-adaptive or continual pretraining. More recent studies have extended this idea to foundation language models. Shi et al. \cite{shi2025continual} describe continual pretraining, domain adaptation, and continual fine-tuning as complementary stages for moving LLMs from general-purpose to specialized capabilities, while \cite{yang2025recent} review continual learning approaches for foundation models, including parameter-efficient and continual pretraining methods.

Supervised fine-tuning (SFT), in contrast, adapts a model using labeled examples designed for a specific task. Recent work suggests that combining domain adaptation with task-specific fine-tuning can be particularly effective. Me-LLaMA \cite{xie2025medical}, for example, combines continued pretraining on large-scale biomedical data with instruction tuning and reports improved performance across multiple medical text analysis tasks. Similarly, Lu et al. \cite{lu2025finetuning} investigate continued pretraining, SFT, and preference optimization for materials science and show that models obtained through different adaptation strategies can provide complementary capabilities when combined. These findings motivate further investigation into how domain and task-specific adaptations can be effectively integrated in specialized LLM applications.

Despite the improvements made when adapting a model to a new domain, further training on domain-specific data can cause catastrophic forgetting, degrading an LLM's existing knowledge and general capabilities \cite{shi2025continual, yang2025recent}. Although parameter-efficient methods such as LoRA can mitigate this issue, they do not eliminate it \cite{biderman2024lora}. Scaling the learned LoRA update thus offers a way to balance domain adaptation with capability preservation by interpolating between the original instruction-tuned model and its fully adapted counterpart, following the principle of weight interpolation \cite{wortsman2022robust}.

\subsection{Natural Language Processing on Oil and Gas Industry}

The oil and gas industry generates large volumes of unstructured technical documentation, particularly through drilling reports. Early studies applied conventional NLP and text-mining methods to these documents \cite{antoniak2016nlp, arumugam2017augmented}, while more recent work has introduced domain-specific transformer models such as PetroBERT \cite{rodrigues2022petrobert} and resources such as Petro NLP \cite{cordeiro2024petronlp} for Portuguese oil-and-gas text.

LLMs have subsequently been explored for petroleum and drilling applications, including information extraction, question answering, and operational decision support \cite{eckroth2023petroqa, asif2024multilabel, pacis2024enhancing, yi2024applications}. More recently, research has focused on adapting LLMs to geoscience and petroleum domains through continued pretraining and parameter-efficient fine-tuning. Examples include K2 \cite{deng2024k2}, GeoGalactica \cite{lin2024geogalactica}, GeoFactory \cite{chen2025geofactory}, EnergyLLM \cite{eckroth2025energyllm}, and QHGeoGPT \cite{ma2026qhgeogpt}, which explore combinations of domain adaptation, LoRA, retrieval, and instruction tuning.

Regardless of these advances, QA over real-world drilling reports using domain-adapted LLMs remains relatively underexplored. Existing studies generally address information extraction, predefined classification tasks, or QA using general-purpose models, while the effect of LoRA adapter scaling on domain and task-specific adaptations has received limited attention. This work addresses these aspects by investigating LoRA-based adaptation for QA over well drilling reports and systematically evaluating the contribution of domain and task adapters at different scaling weights.

\subsection{Parameter-Efficient Fine-Tuning and Model Merging}

Parameter-efficient fine-tuning (PEFT) enables pretrained models to be adapted while keeping most of their original parameters frozen \cite{wang2025peft}. Among PEFT methods, Low-Rank Adaptation (LoRA) \cite{hu2022lora} is widely adopted because it represents parameter updates through low-rank matrices, reducing the computational cost of adaptation while keeping the learned updates separate from the original model weights.

Model merging has explored the possibility of modifying or combining pretrained models after training. Early approaches such as WiSE-FT \cite{wortsman2022robust}, Model Soups \cite{wortsman2022soups}, and Task Arithmetic \cite{ilharco2023editing} showed that model weights or task-specific updates can be interpolated, averaged, or combined to incorporate different capabilities. More recent methods, including TIES-Merging \cite{yadav2023ties} and DARE \cite{yu2024dare}, address interference between model updates, while other studies have investigated weighted combinations of general and domain-adapted models \cite{lu2025finetuning, sasaki2025merging}. These ideas have also been extended to PEFT adapters: AdapterSoup \cite{chronopoulou2023adaptersoup}, LoRAHub \cite{huang2024lorahub}, FashionGPT \cite{gao2024fashiongpt}, and LoRA Soups \cite{prabhakar2025lorasoups} investigate the combination of independently trained adapters, including adaptations targeting different domains or skills. Collectively, these studies show that LoRA adapters can be treated as modular components whose contributions can be adjusted or combined.

However, the role of adapter scaling in domain and task adaptation remains insufficiently explored. In particular, existing studies provide limited systematic analysis of scaling individual LoRA adapters, especially for adapters obtained through DAPT and task-specific fine-tuning. The combination of domain and task-specific LoRA adapters has also received limited attention in specialized technical QA settings. We therefore investigate the effects of scaling individual DAPT and QA adapters, as well as their weighted combination, for question answering over well drilling reports.

\section{Methodology}
\label{sec:method}

\begin{figure*}[t]
  \centering
  \includegraphics[width=\textwidth]{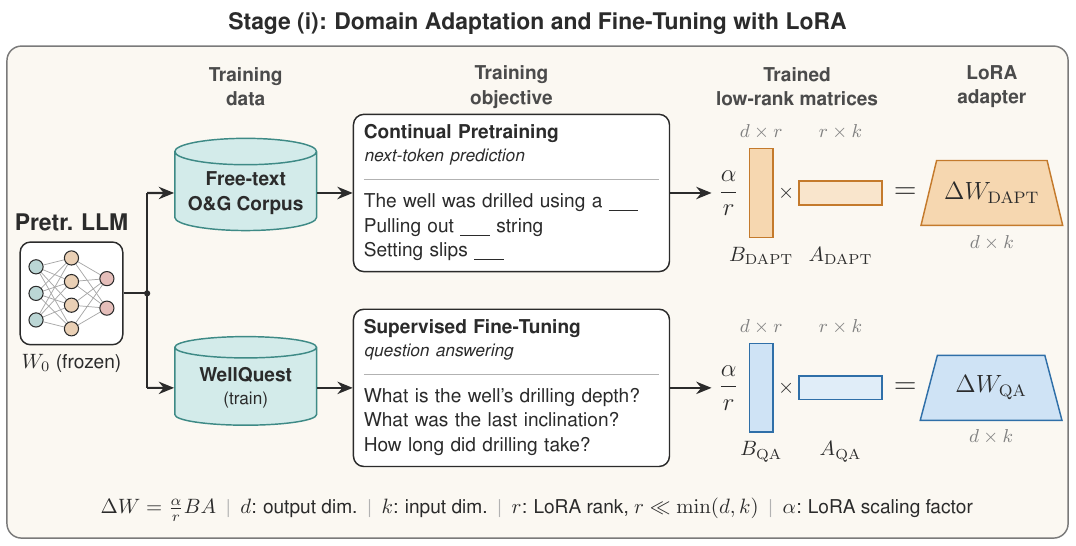}
  \caption{Stage (i) of the methodology. Two LoRA adapters are independently trained from a pretrained LLM with frozen weights $W_0$: the domain adaptation adapter $\Delta W_{\text{DAPT}}$, trained through continued pretraining on unlabeled drilling reports using next-token prediction, and the task-specific adapter $\Delta W_{\text{QA}}$, trained via supervised fine-tuning on the WellQuest training split. Each adapter consists of trainable low-rank matrices $B \in \mathbb{R}^{d \times r}$ and $A \in \mathbb{R}^{r \times k}$, with updates scaled by $\alpha/r$.}
  \label{fig:stage_i}
  \label{fig:pipeline_i}
\end{figure*}

%\begin{figure*}[!htbp]
%        \centering
%        \includegraphics[width=1\linewidth]{figure.tex}
%        \caption{Caption.}        
%        \label{fig:methodology_part2}
%\end{figure*}

\begin{figure*}[t]
  \centering
  \includegraphics[width=\textwidth]{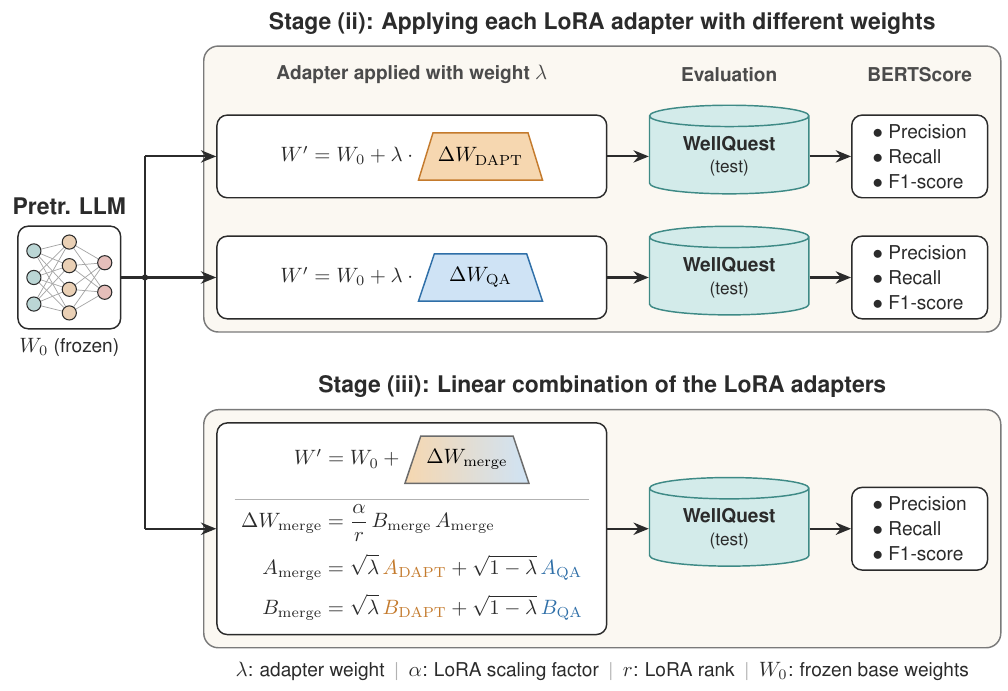}
  \caption{Stages (ii) and (iii) of the methodology. In Stage (ii), each adapter is scaled by $\lambda$ and applied to the frozen model, with performance assessed on the WellQuest test set using BERTScore precision, recall, and F1. In Stage (iii), the DAPT and QA adapters are merged through a linear combination of their low-rank factors, weighted by $\lambda$ and $1 - \lambda$, respectively. The merged model is then evaluated using the same protocol.}
  \label{fig:pipeline_ii_iii}
\end{figure*}

As shown in Figures~\ref{fig:pipeline_i} and~\ref{fig:pipeline_ii_iii}, the methodology comprises three stages. In Stage (i), two LoRA adapters are trained independently: one on unlabeled drilling reports for domain adaptation and the other on QA data through supervised fine-tuning. In Stage (ii), each adapter is applied to the frozen pretrained model at different scaling factors and evaluated on the QA test set. Finally, in Stage (iii), the adapters are linearly
combined using complementary weights, and the resulting models are assessed under the same evaluation protocol.

\subsection{Adapter Training via LoRA}

Starting from a pretrained LLM whose weights $W_0$ remain frozen, we train two LoRA adapters~\cite{hu2022lora}, as illustrated in Figure~\ref{fig:pipeline_i}. LoRA parameterizes the update to each frozen weight matrix $W_0 \in \mathbb{R}^{d \times k}$ as $\Delta W = \frac{\alpha}{r} B A$, where $B \in \mathbb{R}^{d \times r}$ and $A \in \mathbb{R}^{r \times k}$ are trainable low-rank matrices, $k$ and $d$ denote the input and output dimensions, respectively, $r$ is the rank, and $\alpha$ is a scaling factor.

The first adapter, $\Delta W_{\text{DAPT}} = \frac{\alpha}{r} B_{\text{DAPT}} A_{\text{DAPT}}$, is trained through domain-adaptive pretraining (DAPT), using next-token prediction on unlabeled drilling reports to learn domain-specific terminology and writing conventions. The second adapter, $\Delta W_{\text{QA}} = \frac{\alpha}{r} B_{\text{QA}} A_{\text{QA}}$, is trained through supervised fine-tuning on question-answering data containing questions, supporting contexts, and reference answers. Both adapters use the same LoRA configuration, enabling their combination in Stage (iii). Checkpoints are saved at regular training intervals to assess the effects of adapter scaling throughout training.

\subsection{Weighted Application of Individual Adapters}

To assess the impact of adaptation strength on downstream performance, we scale each trained adapter by a coefficient $\lambda$ before applying it to the frozen model (Figure~\ref{fig:pipeline_ii_iii}, Stage (ii)). Given an adapter $\Delta W \in \{\Delta W_{\text{DAPT}}, \Delta W_{\text{QA}}\}$, the resulting model weights are defined as
\begin{equation}
W' = W_0 + \lambda\, \Delta W,
\label{eq:lambda}
\end{equation}
where $\lambda \in \{0.1, 0.2, \dots, 1.0\}$. A value of $\lambda = 1$ applies the adapter at full strength, while $\lambda = 0$ recovers the original model. Intermediate values interpolate between these two configurations, following the principle of weight interpolation between pretrained and fine-tuned models~\cite{wortsman2022robust}. Since scaling is performed after training, different values of $\lambda$ can be evaluated using the same adapter without additional training or data.

For each training checkpoint and scaling factor, the resulting model answers the questions in the QA test set. These responses are then compared with the reference answers using BERTScore~\cite{zhang2020bertscore}, measuring precision, recall, and F1. The evaluation is conducted separately for the DAPT and QA adapters.

\subsection{Complementary Weighted Merging of Domain and Task Adapters}
\label{sec:method_merge}

Finally, to examine the joint contribution of domain and task adaptation, the two adapters are
combined into a single adapter (Figure~\ref{fig:pipeline_ii_iii}, Stage (iii)). The combination is
controlled by a single weight $\lambda \in [0, 1]$: the DAPT adapter receives weight $\lambda$
and the QA adapter receives the complementary weight $1 - \lambda$, so that increasing the
contribution of one adapter reduces the contribution of the other. Since both adapters share
the same rank and scaling factor, they are combined at the level of their low-rank matrices:
\begin{equation}
A_{\text{merge}} = \sqrt{\lambda}\, A_{\text{DAPT}} + \sqrt{1 - \lambda}\, A_{\text{QA}}, \qquad
B_{\text{merge}} = \sqrt{\lambda}\, B_{\text{DAPT}} + \sqrt{1 - \lambda}\, B_{\text{QA}},
\label{eq:merge}
\end{equation}
and the adapted weights become $W' = W_0 + \Delta W_{\text{merge}}$, with
$\Delta W_{\text{merge}} = \frac{\alpha}{r} B_{\text{merge}} A_{\text{merge}}$.

The weight is varied over $\lambda \in \{0.1, 0.2, \dots, 0.9\}$. The boundary cases
$\lambda = 0$ and $\lambda = 1$ are excluded because they correspond to the QA and DAPT
adapters applied alone at full weight, which are already evaluated in Stage (ii). For each
value of $\lambda$, the merged model is also evaluated on the QA test set.

\section{Experimental Setup} \label{sec:setup}

\subsection{DAPT configurations} \label{sec:dapt_config}

Continued pretraining relies on raw, unlabeled text, which differs substantially from the conversational data used to align instruction-tuned models. Applying continued pretraining directly to an instruct model may therefore interfere with its instruction-following capabilities \cite{wu2025shadowft, huang2024chatvector}, leading to catastrophic forgetting. To investigate this issue, we consider two DAPT configurations. In the first, the LoRA adapter is trained directly on the instruction-tuned model. In the second, the adapter is trained on the corresponding base model and subsequently applied to the instruction-tuned model at inference time, following the Shadow-FT approach \cite{wu2025shadowft}. Shadow-FT shows that continued fine-tuning of instruct models can provide limited gains or even degrade performance, while the corresponding base and instruct models have highly similar weights, enabling updates learned on the base model to be transferred to the instruct model. A related strategy is proposed by Chat Vector \cite{huang2024chatvector}, which restores instruction-following capabilities after continued pretraining by incorporating the weight difference between the base and instruction-tuned models. In our setting, however, the transferred update consists of the LoRA adapter learned during DAPT, while supervised QA fine-tuning is always performed directly on the instruction-tuned model.

\subsection{Used Models}

We conduct our experiments with two open-weight model families of similar size: Qwen2.5-7B \cite{qwen2024qwen25} and Llama-3.1-8B \cite{grattafiori2024llama3}. For each family, we use both the pretrained base model and its instruction-tuned counterpart (Qwen2.5-7B-Instruct and Llama-3.1-8B-Instruct). Three factors guided this choice. First, both families officially support Portuguese, the language of all data used in this work. Second, both release base and instruct checkpoints with identical architectures, which is required by the DAPT configuration in which the adapter is trained on the base model and applied to the instruct model (Section~\ref{sec:dapt_config}). Third, using two families from different developers, trained on different data, allows us to assess whether the effects of the adapter weight are consistent across models rather than specific to one of them. All experiments were run on a single NVIDIA H100 GPU.

\subsection{Datasets}

\textbf{Domain-adaptive pretraining corpus.} The DAPT stage uses a private corpus of approximately 17,000 drilling operation reports in Portuguese provided by a corporation in the petroleum industry. These reports consist of free-form and unlabeled descriptions of well drilling operations, covering activity sequences, their durations, and relevant operational events. As they reflect the terminology and writing conventions of drilling documentation, this corpus is representative of the textual characteristics encountered in the downstream QA task.

\textbf{WellQuest}. For supervised fine-tuning and evaluation, we use WellQuest \cite{grimberg2026wellquest}, an open source\footnote{https://data.mendeley.com/datasets/f64ny7mcvz/1} annotated question-answering dataset derived from Daily Drilling Reports (DDRs) from offshore petroleum wells. The dataset contains question–answer pairs in Portuguese, constructed from the DDRs of two offshore drilling interventions, along with the anonymized DDR corpora from both wells. Each instance contains a question, a short answer, an expanded answer, the identifiers of the DDR rows supporting the answer, a contiguous excerpt of activity records providing the relevant context, and classification labels across seven dimensions. WellQuest also provides English translations of its free-text fields, generated through LLM-assisted machine translation and reviewed using an expert-curated domain glossary, as well as a bilingual Portuguese–English glossary of drilling terminology. In this work, we use the original Portuguese version as the authoritative dataset. For our experiments, each instance consists of the question and its corresponding activity context as input, with the expanded answer serving as the target output. We follow the training and test partitions provided by the original dataset.

\subsection{Training Details} \label{sec:training_details}

Both DAPT and QA fine-tuning are performed using QLoRA \cite{dettmers2023qlora}, with the underlying model quantized to 4-bit and kept frozen while LoRA adapters with rank $r=32$ and scaling factor $\alpha=64$ are applied to all attention and feed-forward projection layers. Both training procedures use the AdamW optimizer with a learning rate of $5\times10^{-4}$, a cosine learning-rate schedule, and bfloat16 precision.

For DAPT, the drilling reports are segmented into sequences of 1,024 tokens using a stride corresponding to an overlap of 256 tokens, yielding 94,995 training sequences. Training is performed with an effective batch size of 64 for 1,500 steps, corresponding to slightly more than one pass through the corpus, with checkpoints saved every 300 steps. QA fine-tuning is conducted exclusively on the instruction-tuned models, with the loss calculated only over the answer tokens. Using an effective batch size of 4, training for 300 steps corresponds to approximately 3.7 epochs over the training examples, and checkpoints are saved every 60 steps.

For evaluation, the models are loaded without quantization and generate responses using greedy decoding. The same prompt is used for all configurations, including the baseline. Adapter scaling is implemented by multiplying the scaling factor of each LoRA module by $\lambda$ after loading the adapter, while merged configurations are constructed using the linear combination functionality provided by the PEFT library. BERTScore is computed using the default reference model for Portuguese, i.e., bert-base-multilingual-cased \cite{devlin-etal-2019-bert}.

\section{Results and Discussion} \label{sec:results}

\subsection{Checkpoint analysis} \label{sec:checkpoints}

As the optimal adapter weight may vary throughout training, we evaluate the effect of adapter scaling at multiple training stages. Specifically, every checkpoint saved during DAPT and QA fine-tuning, as defined in Section \ref{sec:training_details}, is evaluated using the complete range of adapter weights. This analysis allows us to select the best checkpoint for each model and training procedure. For DAPT, we evaluate both configurations described in Section \ref{sec:dapt_config}: direct training on the instruction-tuned model and training on the base model followed by transfer to its instruction-tuned counterpart.

\begin{figure*}[!htbp]
        \centering
        \includegraphics[width=1\linewidth]{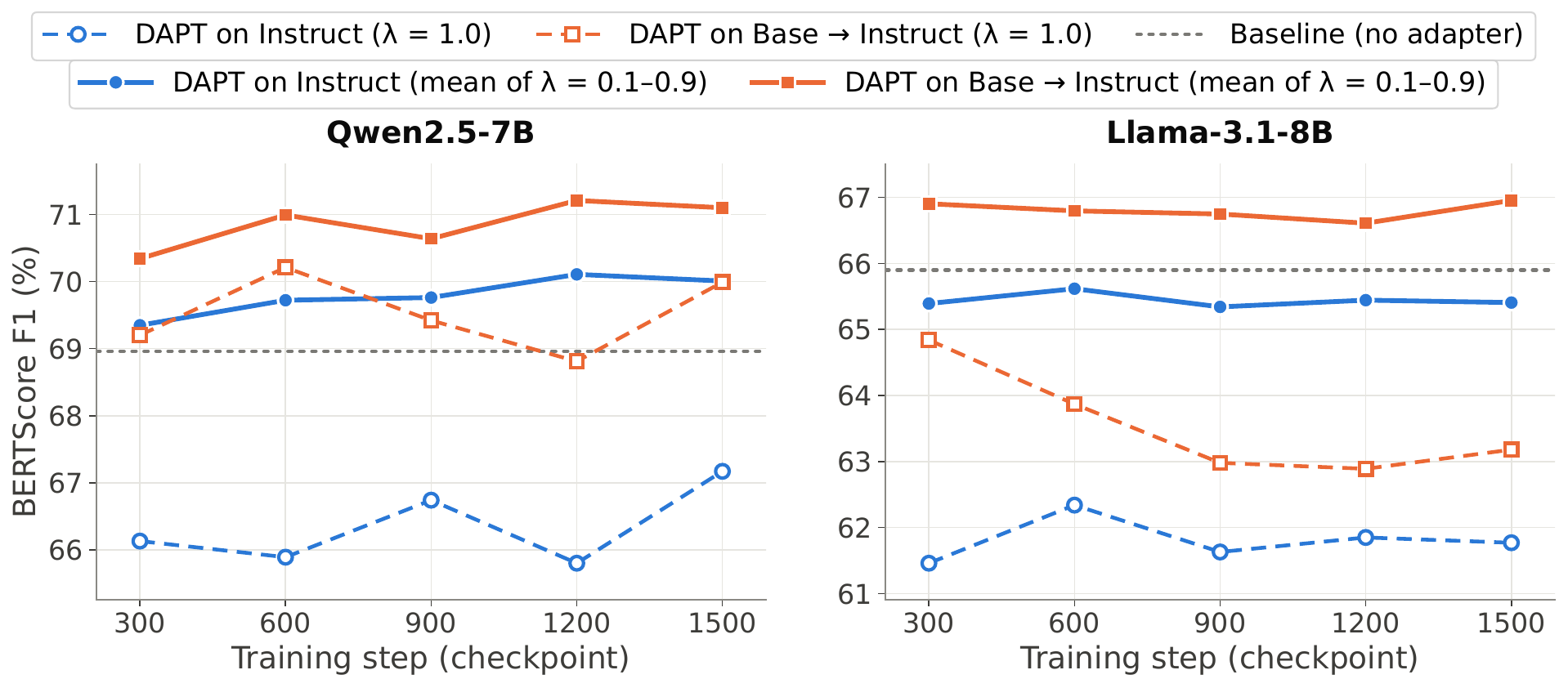}
        \caption{BERTScore F1 on the WellQuest test set across DAPT training checkpoints for Qwen2.5-7B and Llama-3.1-8B. Blue lines represent adapters trained directly on instruction-tuned models, while orange lines represent adapters trained on the corresponding base models and subsequently transferred to the instruction-tuned models. The dotted grey line marks the baseline performance of the instruction-tuned models without adapters.}        
        \label{fig:mean_f1_vs_steps_dapt}
\end{figure*}

Figure \ref{fig:mean_f1_vs_steps_dapt} presents the DAPT results across training steps. For both models and configurations, the mean performance across fractional adapter weights remains above the corresponding full-weight curve at all checkpoints, indicating that applying the DAPT adapter at full strength is consistently less effective. This effect is particularly pronounced when DAPT is performed directly on the instruct model, where the full-weight adapter remains below the baseline throughout training for both models, while the mean over fractional weights exceeds the baseline for Qwen2.5-7B and remains only slightly below it for Llama-3.1-8B. This degradation likely results from a mismatch between the DAPT objective and the instruction-following capabilities acquired during instruction tuning. As domain adaptation primarily optimizes the model for domain-specific language and patterns through continual pretraining with next token prediction, applying it directly to an instruction-tuned model may interfere and degrade its instruction-following behavior and, consequently, reduce it's performance on tasks like QA. Similar interference has been associated with catastrophic forgetting during continued domain training \cite{shi2025continual, yang2025recent} and with degradation of conversational capabilities after fine-tuning on specialized documents \cite{prabhakar2025lorasoups}.

In addition, the benefits of fractional scaling can be understood in terms of the adapted weights. Rather than replacing the original model behavior with a fully DAPT adapted version, a value of $\lambda<1$ interpolates between the instruction-tuned model and the model incorporating the complete DAPT update. Consequently, reducing $\lambda$ allows the model to retain part of the domain-specific information introduced by DAPT while limiting the degradation and forgetting of the  original capabilities of the instruction-tuned model.

Figure~\ref{fig:mean_f1_vs_steps_dapt} also shows a clear advantage of training the DAPT adapter on the base model and subsequently transferring it to the instruct model, compared with training the adapter directly on the instruct model. This advantage is observed for both the mean performance across fractional weights and the full-weight adapter, at every checkpoint and for both models. The difference is particularly pronounced for Llama-3.1-8B: the fractional-weight mean remains below the baseline when DAPT is performed directly on the instruct model, while the transferred adapter consistently remains above it. This behavior is consistent with Shadow-FT \cite{wu2025shadowft}, which argues that base models are more suitable for continued training because their updates can subsequently be transferred to the corresponding instruct models while preserving the instruction-following capabilities acquired during alignment. A related strategy is used by Chat Vector \cite{huang2024chatvector}, which performs continued pretraining on a base model and subsequently transfers the resulting domain knowledge to the instruction-tuned model. Our results suggest that this principle also applies to LoRA-based DAPT on raw technical text, where learning the adapter on the base model avoids directly interfering with the instruction-following behavior of the instruct model. Based on these results, we use the transferred DAPT configuration in the remaining experiments.

Regarding training length, the mean performance across fractional weights of the transferred adapter remains relatively stable across checkpoints, while the full-weight curves exhibit greater variation and, for Llama-3.1, tend to decrease as training progresses. This pattern suggests that increasing the number of DAPT steps may amplify the shift introduced by the domain adaptation, while fractional weighting mitigates this effect and makes performance less dependent on the training checkpoint. Considering the mean performance across fractional LoRA weights under the transferred DAPT configuration, the best results were obtained at step 1200 for Qwen2.5-7B and step 1500 for Llama-3.1-8B, achieving F1 scores of 71.21\% and 66.95\%, respectively. We therefore selected these checkpoints for the subsequent analyses.

\begin{figure*}[!htbp]
        \centering
        \includegraphics[width=1\linewidth]{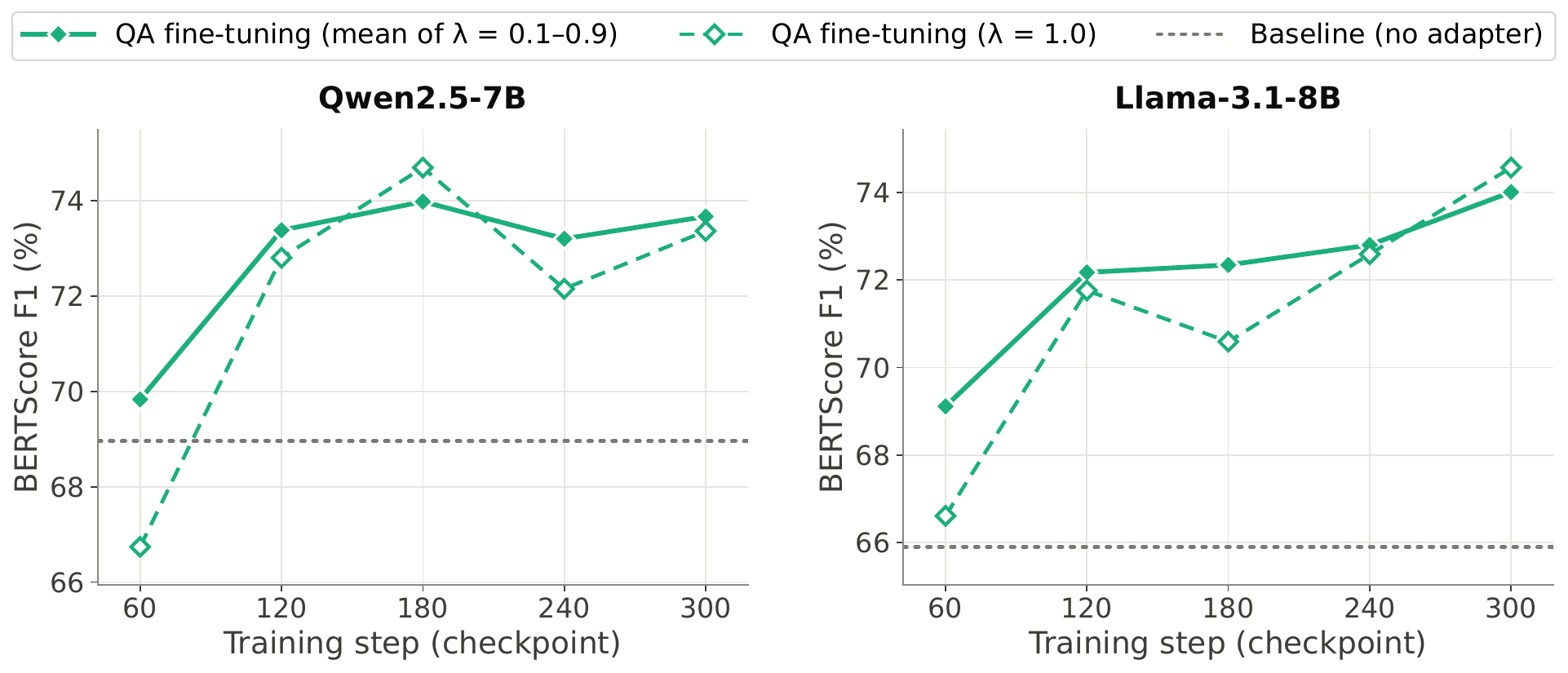}
        \caption{BERTScore F1 on the WellQuest test set across QA fine-tuning training checkpoints for Qwen2.5-7B and Llama-3.1-8B. The dotted grey line marks the baseline performance of the instruction-tuned models without adapters.}        
        \label{fig:mean_f1_vs_steps_qa}
\end{figure*}

Figure~\ref{fig:mean_f1_vs_steps_qa} presents the corresponding analysis for QA fine-tuning. Both models show a sharp improvement between steps 60 and 120 and remain substantially above the baseline as the training continues. Unlike DAPT, however, the relative behavior of fractional and full adapter weights changes throughout training. At the earliest checkpoint, the undertrained adapter performs substantially better at fractional weights, with the full-weight adapter even falling below the baseline for Qwen2.5-7B. As training progresses, this difference decreases, and at step 180 for Qwen2.5-7B and step 300 for Llama-3.1-8B the full-weight adapter achieves higher performance than the fractional-weight mean. It is important to note that a lower mean performance across fractional weights does not imply that every fractional weight performs worse than the full-weight setting.
Despite this, it is interesting to note that both the full-weight adapter and the mean performance across fractional weights reach their highest values at step 180 for Qwen2.5-7B and step 300 for Llama-3.1-8B. At step 180, Qwen2.5-7B achieves F1 scores of 74.69\% with the full-weight adapter and 73.98\% averaged across fractional weights. For Llama-3.1-8B at step 300, the corresponding scores are 74.57\% and 74.01\%, respectively. Based on these results, we select step 180 for Qwen2.5-7B and step 300 for Llama-3.1-8B on QA fine-tuning for the subsequent analyses.

\subsection{Effect of the Adapter Weight $\lambda$} \label{sec:lambda}

Using the checkpoints selected in Section \ref{sec:checkpoints}, we next investigate the effect of the adapter scaling factor $\lambda$ on QA performance. For a given trained adapter, varying $\lambda$ changes only the magnitude of its low-rank update, allowing different adapted models to be evaluated without additional training. The conventional setting $\lambda=1.0$ applies the adapter at its full strength, whereas values below 1.0 reduce its contribution and interpolate between the original instruct model and the fully adapted model.

\begin{figure*}[!htbp]
        \centering
        \includegraphics[width=1\linewidth]{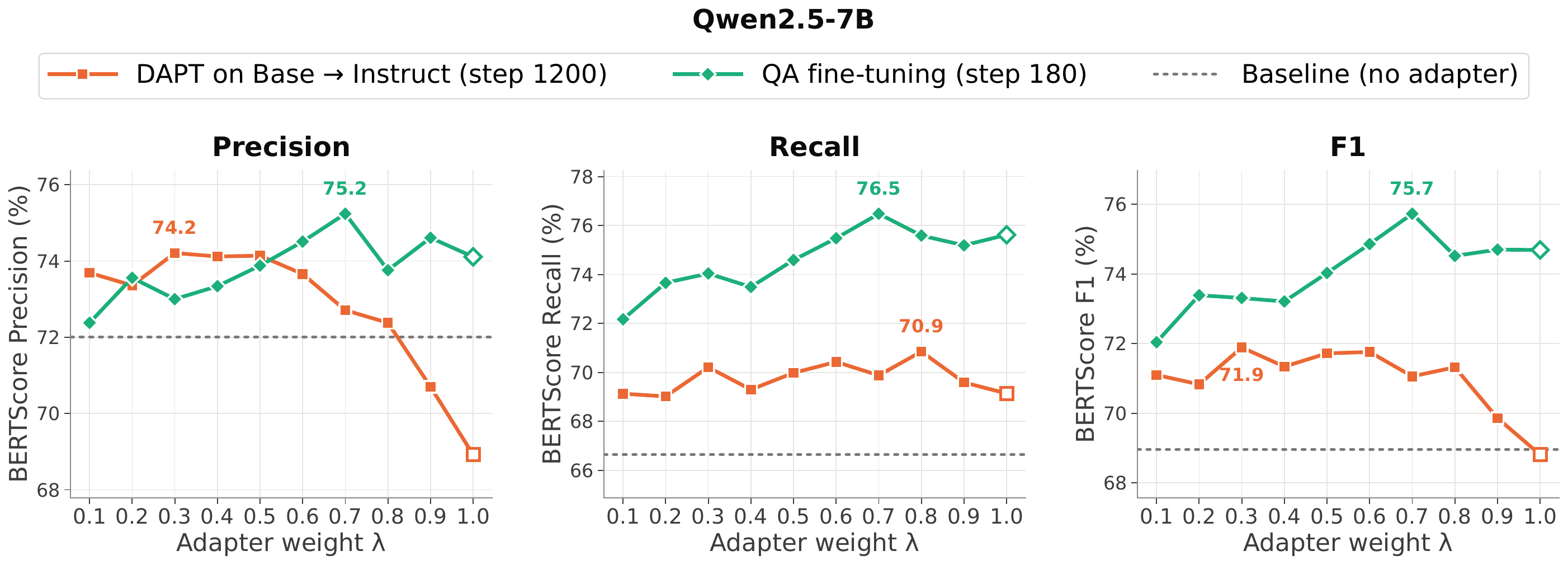}
        \caption{BERTScore precision, recall, and F1 for Qwen2.5-7B on the WellQuest test set across adapter scaling factors ($\lambda$), for the base model trained DAPT adapter transferred to the instruction-tuned model (step 1200) and the QA adapter (step 180). Hollow markers denote full-weight adapters ($\lambda = 1.0$). Annotated values identify the optimal scaling factor for each curve and the dotted line represents the unadapted instruction-tuned model baseline.}        
        \label{fig:metrics_vs_lambda_qwen}
\end{figure*}

Figure \ref{fig:metrics_vs_lambda_qwen} reports BERTScore precision, recall, and F1 as a function of $\lambda$ for the DAPT and QA adapters on the Qwen2.5-7B model considering the training steps 1200 and 180 for DAPT and QA, respectively. In the case of the DAPT adapter, precision and F1 peak at $\lambda = 0.3$, with an F1 of 71.9\%, remaining relatively stable up to $\lambda = 0.8$, and drop at the two largest weights. At $\lambda = 1.0$, precision falls below the baseline and F1 returns to the baseline level, so the gain obtained with intermediate weights disappears when the adapter is fully applied. Recall, on other hand, behaves differently, staying above the baseline for all weights and changes little with $\lambda$. However, it is still worth noting that several intermediate $\lambda$ values achieve better recall than $\lambda = 1.0$, with $\lambda = 0.8$ yielding the best result.

The QA adapter on Qwen2.5-7B improves on the baseline across the three metrics, with the clearest gains in recall and F1. Its curves rise with $\lambda$ and reach their maximum at $\lambda = 0.7$ in all three metrics, with an F1 of 75.7\%, followed by a drop at $\lambda = 0.8$ and a slightly lower plateau up to $\lambda = 1.0$. Overall, the QA adapter outperforms the DAPT adapter across most $\lambda$ values. This result is expected because the downstream task is question answering, and the QA adapter is directly optimized for this objective through supervised fine-tuning. In contrast, DAPT is trained on unlabeled domain-specific text through next token prediction and primarily captures the terminology, linguistic patterns, and characteristics of the target domain without optimizing the model for question answering. Therefore, although DAPT provides domain-specific knowledge that can support downstream tasks, the QA adapter is better aligned with the objective evaluated in this study.

\begin{figure*}[!htbp]
        \centering
        \includegraphics[width=1\linewidth]{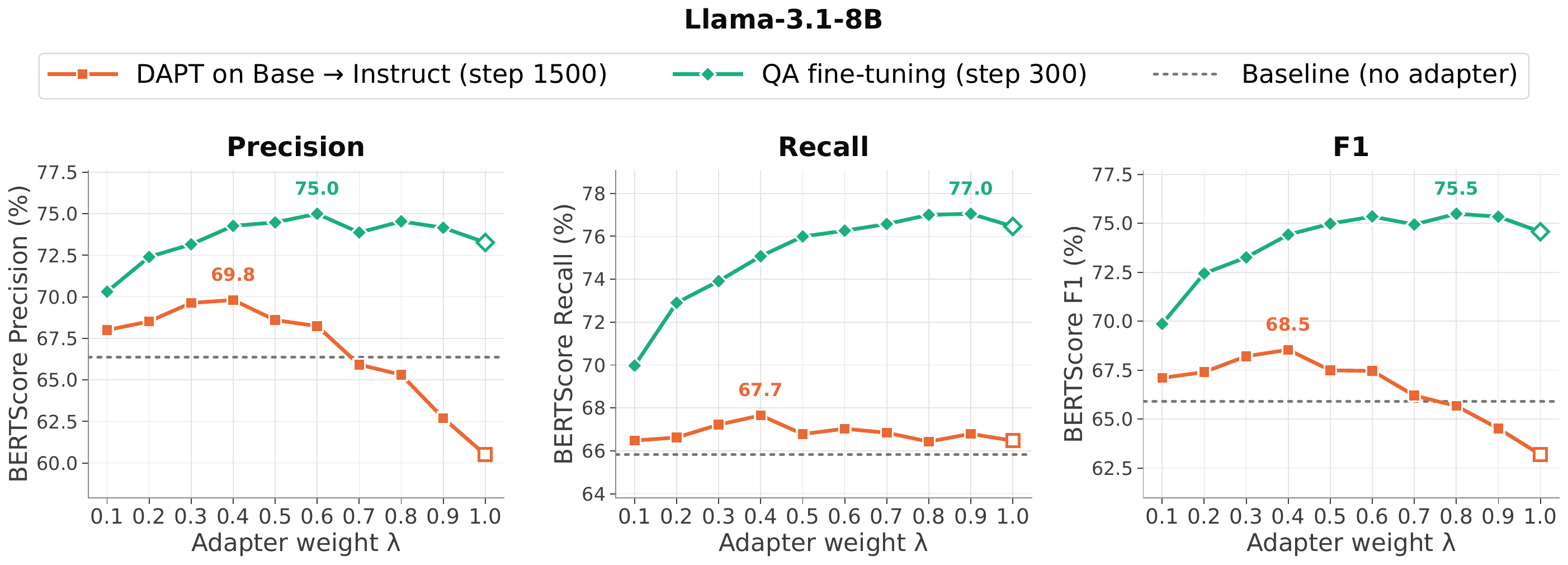}
        \caption{BERTScore precision, recall, and F1 for Llama-3.1-8B on the WellQuest test set across adapter scaling factors ($\lambda$), for the base model trained DAPT adapter transferred to the instruction-tuned model (step 1500) and the QA adapter (step 300). Hollow markers denote full-weight adapters ($\lambda = 1.0$). Annotated values identify the optimal scaling factor for each curve and the dotted line represents the unadapted instruction-tuned model baseline.}        
        \label{fig:metrics_vs_lambda_llama}
\end{figure*}

Figure~\ref{fig:metrics_vs_lambda_llama} presents the BERTScore results obtained with different $\lambda$ values for the Llama-3.1-8B model considering the training step 1500 for DAPT and 300 for QA fine-tuning. For the DAPT adapter, the overall behavior is similar to that observed for Qwen2.5-7B. Precision, recall, and F1 reach their highest values at $\lambda=0.4$, with F1 reaching 68.5\%, after which performance declines progressively, particularly for precision and F1. Precision falls below the baseline from $\lambda=0.7$ onward, causing F1 to also drop below the baseline at the largest weights. In contrast, once again, recall remains above the baseline and relatively stable across the range. 

In this case, the QA adapter also consistently outperforms the DAPT adapter across all metrics and scaling factors. Its performance increases substantially from $\lambda=0.1$ to $\lambda=0.5$ and then remains relatively stable, with the highest F1 obtained at $\lambda=0.8$. Within this range, precision reaches its maximum earlier, at $\lambda=0.6$, while recall continues to improve toward the largest weights. Nevertheless, as observed for Qwen2.5-7B, the results show that applying the QA adapter at full strength does not provide the best overall performance, further supporting the benefit of controlling the adapter contribution through $\lambda$.

Despite differences in magnitude, both models exhibit the same overall pattern. The DAPT adapter performs best at low to intermediate weights and deteriorates at full weight, with the degradation primarily affecting precision while recall remains relatively stable. This behavior is consistent with the mismatch between the DAPT objective and the instruction-response format required for QA, as discussed in Section \ref{sec:checkpoints}. Therefore, reducing the adapter weight can preserve part of the domain-specific signal while limiting the shift away from the original instruction-following behavior, consistent with the benefits of weight interpolation reported in previous work \cite{ wortsman2022robust}. In contrast, the QA adapter performs best at relatively high weights, reflecting its direct optimization for the target QA task, although its performance tends to plateau and deteriorate in the upper range rather than consistently favoring $\lambda=1.0$. Thus, as the two adapters exhibit complementary scaling behavior, with DAPT favoring lower weights and QA favoring higher ones, we decided to investigate the complementary combination of the DAPT and QA LoRA adapters to determine whether jointly incorporating their domain and task-specific contributions can further improve performance.

\subsection{Merging Adapters}

After evaluating the two adapters independently, we next examine whether their combination can further improve performance relative to the QA adapter alone, which achieved the strongest overall results. Using the formulation introduced in Section~\ref{sec:method_merge}, we combine the DAPT and QA adapters with complementary weights based on the checkpoints selected in Section~\ref{sec:checkpoints}. This linear interpolation of independently learned updates is related to task arithmetic and adapter averaging \cite{ilharco2023editing, chronopoulou2023adaptersoup}, and more closely follows the complementary interpolation strategy proposed by Sasaki et al. \cite{sasaki2025merging}. The endpoints of the interpolation range correspond to configurations evaluated previously: $\lambda = 0$ yields the QA adapter at full weight, while $\lambda = 1$ yields the DAPT adapter at full weight.

\begin{figure*}[!htbp]
        \centering
        \includegraphics[width=1\linewidth]{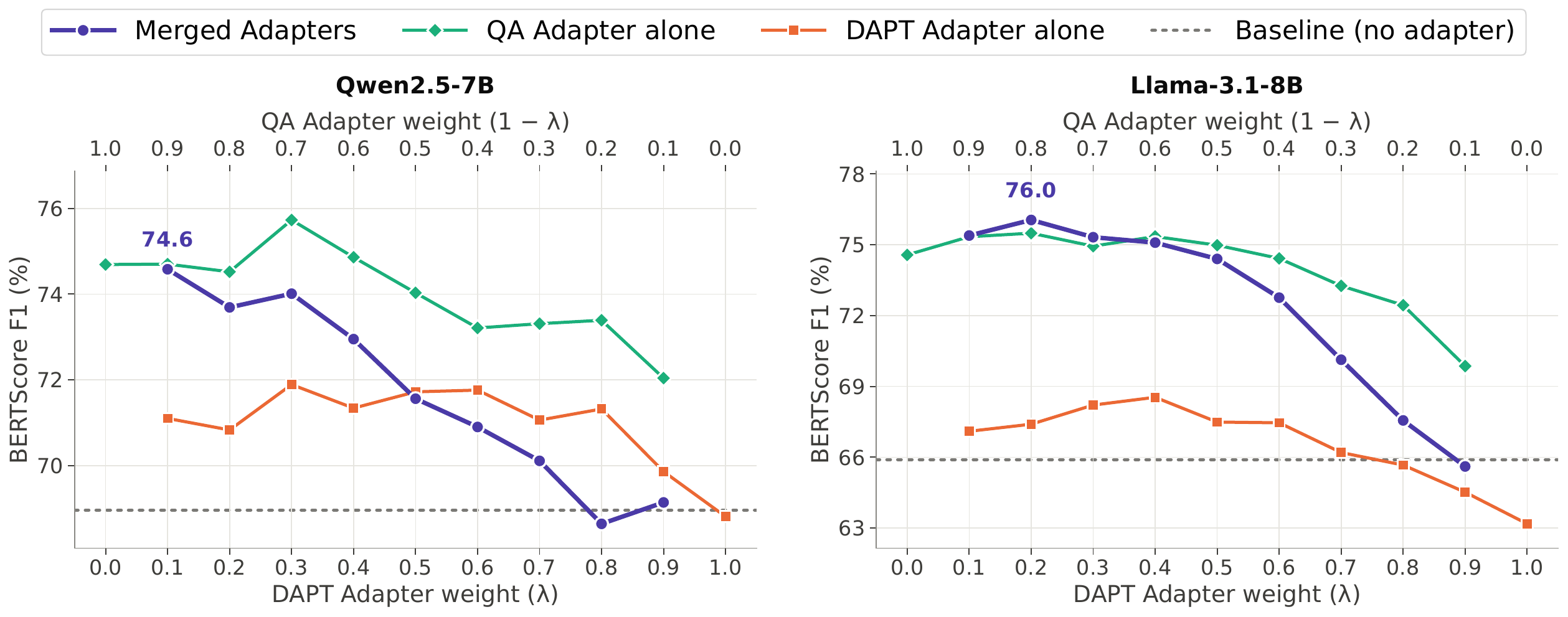}
        \caption{BERTScore F1 on the WellQuest test set for merged DAPT and QA adapters in Qwen2.5-7B and Llama-3.1-8B, plotted as a function of the DAPT weight $\lambda$ (bottom axis) and the complementary QA weight $1-\lambda$ (top axis). Merging is evaluated for $\lambda \in \{0.1,..., 0.9\}$, with each adapter's standalone performance also reported at its corresponding merge weight. The annotated value identifies the best merged configuration, while the dotted line represents the unadapted instruction-tuned model baseline.}        
        \label{fig:merge_adapters}
\end{figure*}

Figure~\ref{fig:merge_adapters} shows the F1 scores obtained by the merged models as a function of the DAPT adapter weight $\lambda$ (bottom axis), while the corresponding weight assigned to the QA adapter, $1-\lambda$, is indicated on the top axis. For each value of $\lambda$, the figure also reports the performance of the DAPT and QA adapters when applied individually at their respective merge weights. This comparison allows us to distinguish the effect of combining the two adapters from the effect of simply scaling an individual one.

For Qwen2.5-7B, the highest F1 among the merged configurations is obtained at $\lambda = 0.1$, reaching 74.6\%. This result is nearly identical to that of the QA adapter applied alone with a weight $1 - \lambda = 0.9$ and to the full-weight QA adapter. Nevertheless, it remains below the best QA configuration reported in Section~\ref{sec:lambda}. As the scaling factor of the DAPT adapter increases, the performance of the merged model progressively diverges from the QA-alone curve, indicating that incorporating the DAPT update does not provide an additional benefit at any of the evaluated weights. For $\lambda \geq 0.5$, the merged model also underperforms the DAPT adapter applied alone at the corresponding weight, indicating that in this range the combination is inferior to either adapter individually. At $\lambda= 0.8$, the F1 of the merged model reaches approximately the baseline level and remains close to it through $\lambda = 0.9$.

In contrast, Llama-3.1-8B exhibits a more favorable response to small contributions from the DAPT adapter. At $\lambda = 0.1$, the merged model achieves essentially the same F1 as the QA adapter alone, while further increasing the DAPT contribution to $\lambda = 0.2$ and $\lambda = 0.3$ improves performance beyond the QA-alone configuration. The best result is obtained at $\lambda = 0.2$, with an F1 of 76.0\%, which represents the highest score observed for this model and slightly exceeds the best QA configuration reported in Section~\ref{sec:lambda}. Beyond $\lambda = 0.3$, however, the performance of the merged model progressively declines relative to the QA curve, with the difference becoming more pronounced for $\lambda > 0.5$. At $\lambda = 0.9$, the merged model reaches approximately the baseline performance. Despite this decline at higher DAPT weights, the merged configuration remains above the DAPT-alone curve across the entire range of $\lambda$, indicating that the QA adapter consistently provides a positive contribution when combined with DAPT on the Llama-3.1-8B model.

Overall, both models achieve their best merged performance with a small DAPT contribution ($\lambda= 0.1$ for Qwen2.5-7B and $\lambda = 0.2$ for Llama-3.1-8B), consistent with the complementary scaling behavior observed in Section~\ref{sec:lambda}, where DAPT favors lower weights and QA favors higher ones. However, the effect of combining the adapters is model-dependent. Llama-3.1-8B obtains a modest improvement over its best QA configuration, whereas Qwen2.5-7B does not benefit from the combination and performs below both individual adapters at intermediate and high DAPT weights. This behavior may reflect interference between the independently trained updates, a known challenge in simple parameter combination \cite{yadav2023ties}. Overall, these results suggest that combining DAPT and QA adapters can provide gains in some models, but its effectiveness should be evaluated empirically rather than assumed to be universally beneficial.

%\section{\textcolor{red}{Responses Qualitatitve Analysis}}

%\textcolor{red}{Nessa seção eu pretendo pegar os top 2 modelos que deram os melhores F1 e comparar a respostas deles com a respostas originais e dos modelos baselines. Se caber na tabela, seria legal colocar as respostas dos modelos grandes como GPT, Claude, etc...}

\section{Conclusion} \label{sec:conclusion}

This work examined the effect of LoRA adapter scaling on question answering over Portuguese well drilling reports using Qwen2.5-7B and Llama-3.1-8B. Three adaptation strategies were evaluated: DAPT on a private drilling-report corpus, either directly on instruction-tuned models or through base-to-instruct transfer; supervised fine-tuning on WellQuest; and weighted merging of DAPT and QA adapters. Scaling factors from 0.1 to 1.0 were tested across multiple training checkpoints, with performance assessed using BERTScore precision, recall, and F1 against full-weight adapters and the unadapted instruction-tuned models.

The results indicate that full-weight LoRA adapters are not always optimal. For DAPT, fractional scaling consistently outperformed full-weight adaptation, increasing F1 from 68.8\% to 71.9\% for Qwen2.5-7B and from 63.2\% to 68.5\% for Llama-3.1-8B. QA fine-tuning also benefited from fractional scaling, although with smaller gains. Moreover, transferring DAPT adapters trained on base models to their instruction-tuned counterparts outperformed direct training, extending the Shadow-FT \cite{wu2025shadowft} strategy to LoRA-based DAPT in the oil and gas domain. Finally, weighted adapter merging yielded a modest F1 improvement for Llama-3.1-8B (76.0\% versus 75.5\%) but no gain for Qwen2.5-7B, indicating that its effectiveness is model dependent.

From a practical perspective, adapter scaling offers a low-cost way to improve LLM adaptation without additional data or training, particularly in specialized domains with limited data and costly expert annotation. Since the effects vary across adapter types and models, it is advisable to treat the scale factor as a hyperparameter for optimizing model performance instead of fixing the adapter at 1.0 weight.

Despite the contributions, this study has some limitations, including the use of a single dataset and language, and BERTScore as a metric, which measures only semantic similarity to the reference answers, and not factual accuracy. The observed precision degradation under full-weight DAPT suggests potential catastrophic forgetting, but its impact on general capabilities remains unexplored. Additionally, the linear adapter merge of the PEFT library may introduce interference between low-rank updates, since this combination calculates cross terms between the adapters. In this sense, for future work, we intend to explore broader models, languages, and domains; assess factual accuracy through human or LLM-based evaluations and compare our method with alternative adapter-merging methods.

\section*{Acknowledgments}

 L. C. Ribas acknowledges support from São Paulo Research Foundation (FAPESP) (\#2025/14029-8 and \#2025/27044-5). The authors would like to thank the Brazilian National Council for Scientific and Technological Development (CNPq) for the financial support (130337/2026-9). The authors would like to thank Petrobras (grant \#3681/2024-CCP)
The authors gratefully acknowledge the support of the Coordenação de 
Aperfeiçoamento de Pessoal de Nível Superior -- Brasil (CAPES).

 % \section*{Author contributions}

 % \textcolor{red}{Completar essa seção...}

 \bibliography{references}

@misc{qwen2024qwen25,
  title         = {{Qwen2.5} Technical Report},
  author        = {{Qwen Team}},
  year          = {2024},
  eprint        = {2412.15115},
  archivePrefix = {arXiv},
  primaryClass  = {cs.CL}
}

@misc{grattafiori2024llama3,
  title         = {The {Llama} 3 Herd of Models},
  author        = {Grattafiori, Aaron and others},
  year          = {2024},
  eprint        = {2407.21783},
  archivePrefix = {arXiv},
  primaryClass  = {cs.AI},
}

@inproceedings{zhang2020bertscore,
  title     = {{BERTScore}: Evaluating Text Generation with {BERT}},
  author    = {Zhang, Tianyi and Kishore, Varsha and Wu, Felix and Weinberger, Kilian Q. and Artzi, Yoav},
  booktitle = {International Conference on Learning Representations (ICLR)},
  year      = {2020}
}

@misc{grimberg2026wellquest,
  title        = {{WellQuest}},
  author       = {Grimberg, Marcelo and Paes, Diogo Melo and Lima, Claudio Benevenuto de Campos and Policarpo, Nara Ang{\'e}lica and Sampaio Pinto, Marcio Augusto},
  year         = {2026},
  howpublished = {Mendeley Data, Version 1},
  doi          = {10.17632/f64ny7mcvz.1}
}

@inproceedings{gururangan2020dont,
  title     = {Don't Stop Pretraining: Adapt Language Models to Domains and Tasks},
  author    = {Gururangan, Suchin and Marasovi{\'c}, Ana and Swayamdipta, Swabha and Lo, Kyle and Beltagy, Iz and Downey, Doug and Smith, Noah A.},
  booktitle = {Proceedings of the 58th Annual Meeting of the Association for Computational Linguistics (ACL)},
  pages     = {8342--8360},
  year      = {2020},
  doi       = {10.18653/v1/2020.acl-main.740}
}

@article{shi2025continual,
  title   = {Continual Learning of Large Language Models: A Comprehensive Survey},
  author  = {Shi, Haizhou and Xu, Zihao and Wang, Hengyi and Qin, Weiyi and Wang, Wenyuan and Wang, Yibin and Wang, Zifeng and Ebrahimi, Sayna and Wang, Hao},
  journal = {ACM Computing Surveys},
  volume  = {58},
  number  = {5},
  pages   = {120:1--120:42},
  year    = {2025},
  doi     = {10.1145/3735633}
}

@article{yang2025recent,
  title   = {Recent Advances of Foundation Language Models-based Continual Learning: A Survey},
  author  = {Yang, Yutao and Zhou, Jie and Ding, Xuanwen and Huai, Tianyu and Liu, Shunyu and Chen, Qin and Xie, Yuan and He, Liang},
  journal = {ACM Computing Surveys},
  volume  = {57},
  number  = {5},
  pages   = {112:1--112:38},
  year    = {2025},
  doi     = {10.1145/3705725}
}

@article{xie2025medical,
  title   = {Medical Foundation Large Language Models for Comprehensive Text Analysis and Beyond},
  author  = {Xie, Qianqian and Chen, Qingyu and Chen, Aokun and Peng, Cheng and Hu, Yan and Lin, Fongci and Peng, Xueqing and Huang, Jimin and Zhang, Jeffrey and Keloth, Vipina and Zhou, Xinyu and Qian, Lingfei and He, Huan and Shung, Dennis and Ohno-Machado, Lucila and Wu, Yonghui and Xu, Hua and Bian, Jiang},
  journal = {npj Digital Medicine},
  volume  = {8},
  pages   = {141},
  year    = {2025},
  doi     = {10.1038/s41746-025-01533-1}
}

@article{lu2025finetuning,
  title   = {Fine-Tuning Large Language Models for Domain Adaptation: Exploration of Training Strategies, Scaling, Model Merging and Synergistic Capabilities},
  author  = {Lu, Wei and Luu, Rachel K. and Buehler, Markus J.},
  journal = {npj Computational Materials},
  volume  = {11},
  pages   = {84},
  year    = {2025},
  doi     = {10.1038/s41524-025-01564-y}
}

@misc{wu2025shadowft,
  title         = {{Shadow-FT}: Tuning Instruct Model via Training on Paired Base Model},
  author        = {Wu, Taiqiang and Yang, Runming and Li, Jiayi and Hu, Pengfei and Wu, Yik-Chung and Wong, Ngai and Yang, Yujiu},
  year          = {2025},
  eprint        = {2505.12716},
  archivePrefix = {arXiv},
  primaryClass  = {cs.CL},
  note          = {Preprint}
}

@inproceedings{huang2024chatvector,
  title     = {Chat Vector: A Simple Approach to Equip {LLMs} with Instruction Following and Model Alignment in New Languages},
  author    = {Huang, Shih-Cheng and Li, Pin-Zu and Hsu, Yu-Chi and Chen, Kuang-Ming and Lin, Yu Tung and Hsiao, Shih-Kai and Tsai, Richard Tzong-Han and Lee, Hung-yi},
  booktitle = {Proceedings of the 62nd Annual Meeting of the Association for Computational Linguistics (Volume 1: Long Papers)},
  year      = {2024},
  doi       = {10.18653/v1/2024.acl-long.590}
}

@inproceedings{hu2022lora,
  title     = {{LoRA}: Low-Rank Adaptation of Large Language Models},
  author    = {Hu, Edward J. and Shen, Yelong and Wallis, Phillip and Allen-Zhu, Zeyuan and Li, Yuanzhi and Wang, Shean and Wang, Lu and Chen, Weizhu},
  booktitle = {International Conference on Learning Representations (ICLR)},
  year      = {2022}
}

@article{wang2025peft,
  title   = {Parameter-Efficient Fine-Tuning in Large Language Models: A Survey of Methodologies},
  author  = {Wang, Luping and Chen, Sheng and Jiang, Linnan and Pan, Shu and Cai, Runze and Yang, Sen and Yang, Fei},
  journal = {Artificial Intelligence Review},
  volume  = {58},
  pages   = {227},
  year    = {2025},
  doi     = {10.1007/s10462-025-11236-4}
}

@article{biderman2024lora,
  title   = {{LoRA} Learns Less and Forgets Less},
  author  = {Biderman, Dan and Portes, Jacob and Gonzalez Ortiz, Jose Javier and Paul, Mansheej and Greengard, Philip and Jennings, Connor and King, Daniel and Havens, Sam and Chiley, Vitaliy and Frankle, Jonathan and Blakeney, Cody and Cunningham, John P.},
  journal = {Transactions on Machine Learning Research},
  year    = {2024}
}

@inproceedings{zhang2025uora,
  title     = {{UORA}: Uniform Orthogonal Reinitialization Adaptation in Parameter Efficient Fine-Tuning of Large Models},
  author    = {Zhang, Xueyan and Zhao, Jinman and Yang, Zhifei and Zhong, Yibo and Guan, Shuhao and Cao, Linbo and Wang, Yining},
  booktitle = {Proceedings of the 63rd Annual Meeting of the Association for Computational Linguistics (Volume 1: Long Papers)},
  pages     = {11709--11728},
  year      = {2025}
}

@inproceedings{wortsman2022robust,
  title     = {Robust Fine-Tuning of Zero-Shot Models},
  author    = {Wortsman, Mitchell and Ilharco, Gabriel and Kim, Jong Wook and Li, Mike and Kornblith, Simon and Roelofs, Rebecca and Gontijo Lopes, Raphael and Hajishirzi, Hannaneh and Farhadi, Ali and Namkoong, Hongseok and Schmidt, Ludwig},
  booktitle = {Proceedings of the IEEE/CVF Conference on Computer Vision and Pattern Recognition (CVPR)},
  year      = {2022}
}

@inproceedings{wortsman2022soups,
  title     = {Model Soups: Averaging Weights of Multiple Fine-Tuned Models Improves Accuracy without Increasing Inference Time},
  author    = {Wortsman, Mitchell and Ilharco, Gabriel and Gadre, Samir Yitzhak and Roelofs, Rebecca and Gontijo-Lopes, Raphael and Morcos, Ari S. and Namkoong, Hongseok and Farhadi, Ali and Carmon, Yair and Kornblith, Simon and Schmidt, Ludwig},
  booktitle = {Proceedings of the 39th International Conference on Machine Learning (ICML)},
  series    = {Proceedings of Machine Learning Research},
  volume    = {162},
  year      = {2022}
}

@inproceedings{ilharco2023editing,
  title     = {Editing Models with Task Arithmetic},
  author    = {Ilharco, Gabriel and Ribeiro, Marco Tulio and Wortsman, Mitchell and Gururangan, Suchin and Schmidt, Ludwig and Hajishirzi, Hannaneh and Farhadi, Ali},
  booktitle = {International Conference on Learning Representations (ICLR)},
  year      = {2023}
}

@inproceedings{yadav2023ties,
  title     = {{TIES-Merging}: Resolving Interference When Merging Models},
  author    = {Yadav, Prateek and Tam, Derek and Choshen, Leshem and Raffel, Colin and Bansal, Mohit},
  booktitle = {Advances in Neural Information Processing Systems (NeurIPS)},
  volume    = {36},
  year      = {2023}
}

@inproceedings{yu2024dare,
  title     = {Language Models are Super Mario: Absorbing Abilities from Homologous Models as a Free Lunch},
  author    = {Yu, Le and Yu, Bowen and Yu, Haiyang and Huang, Fei and Li, Yongbin},
  booktitle = {Proceedings of the 41st International Conference on Machine Learning (ICML)},
  series    = {Proceedings of Machine Learning Research},
  volume    = {235},
  year      = {2024}
}

@inproceedings{chronopoulou2023adaptersoup,
  title     = {{AdapterSoup}: Weight Averaging to Improve Generalization of Pretrained Language Models},
  author    = {Chronopoulou, Alexandra and Peters, Matthew and Fraser, Alexander and Dodge, Jesse},
  booktitle = {Findings of the Association for Computational Linguistics: EACL 2023},
  year      = {2023},
  doi       = {10.18653/v1/2023.findings-eacl.153}
}

@inproceedings{huang2024lorahub,
  title     = {{LoraHub}: Efficient Cross-Task Generalization via Dynamic {LoRA} Composition},
  author    = {Huang, Chengsong and Liu, Qian and Lin, Bill Yuchen and Pang, Tianyu and Du, Chao and Lin, Min},
  booktitle = {Conference on Language Modeling (COLM)},
  year      = {2024}
}

@article{gao2024fashiongpt,
  title   = {{FashionGPT}: {LLM} Instruction Fine-Tuning with Multiple {LoRA}-Adapter Fusion},
  author  = {Gao, Dehong and Ma, Yufei and Liu, Sen and Song, Mengfei and Jin, Linbo and Jiang, Wen and Wang, Xin and Ning, Wei and Yu, Shanqing and Xuan, Qi and Cai, Xiaoyan and Yang, Libin},
  journal = {Knowledge-Based Systems},
  volume  = {299},
  pages   = {112043},
  year    = {2024},
  doi     = {10.1016/j.knosys.2024.112043}
}

@inproceedings{prabhakar2025lorasoups,
  title     = {{LoRA} Soups: Merging {LoRAs} for Practical Skill Composition Tasks},
  author    = {Prabhakar, Akshara and Li, Yuanzhi and Narasimhan, Karthik and Kakade, Sham and Malach, Eran and Jelassi, Samy},
  booktitle = {Proceedings of the 31st International Conference on Computational Linguistics: Industry Track (COLING)},
  pages     = {644--655},
  year      = {2025}
}

@inproceedings{sasaki2025merging,
  title     = {Effect of Model Merging in Domain-Specific Ad-hoc Retrieval},
  author    = {Sasaki, Taiga and Yamamoto, Takehiro and Ohshima, Hiroaki and Fujita, Sumio},
  booktitle = {Proceedings of the 34th ACM International Conference on Information and Knowledge Management (CIKM)},
  pages     = {5208--5212},
  year      = {2025},
  doi       = {10.1145/3746252.3760920}
}

@inproceedings{rodrigues2022petrobert,
  title     = {{PetroBERT}: A Domain Adaptation Language Model for Oil and Gas Applications in {Portuguese}},
  author    = {Rodrigues, Rafael B. M. and Privatto, Pedro I. M. and de Sousa, Gustavo Jos{\'e} and Murari, Rafael P. and Afonso, Luis C. S. and Papa, Jo{\~a}o P. and Pedronette, Daniel C. G. and Guilherme, Ivan R. and Perrout, Stephan R. and Riente, Aliel F.},
  booktitle = {Computational Processing of the Portuguese Language (PROPOR 2022)},
  series    = {Lecture Notes in Computer Science},
  volume    = {13208},
  pages     = {101--109},
  publisher = {Springer},
  year      = {2022},
}

@article{cordeiro2024petronlp,
  title   = {{Petro NLP}: Resources for Natural Language Processing and Information Extraction for the Oil and Gas Industry},
  author  = {Cordeiro, F{\'a}bio Corr{\^e}a and da Silva, Patr{\'i}cia Ferreira and Tessarollo, Alexandre and Freitas, Cl{\'a}udia and de Souza, Elvis and Gomes, Diogo da Silva Magalh{\~a}es and Souza, Renato Rocha and Coelho, Fl{\'a}vio Code{\c{c}}o},
  journal = {Computers \& Geosciences},
  volume  = {193},
  pages   = {105714},
  year    = {2024},
  doi     = {10.1016/j.cageo.2024.105714}
}

@article{ma2024information,
  title   = {Information Extraction from Historical Well Records Using a Large Language Model},
  author  = {Ma, Zhiwei and Santos, Javier E. and Lackey, Greg and Viswanathan, Hari and O'Malley, Daniel},
  journal = {Scientific Reports},
  volume  = {14},
  pages   = {31702},
  year    = {2024},
  doi     = {10.1038/s41598-024-81846-5}
}

@inproceedings{deng2024k2,
  title     = {{K2}: A Foundation Language Model for Geoscience Knowledge Understanding and Utilization},
  author    = {Deng, Cheng and Zhang, Tianhang and He, Zhongmou and Chen, Qiyuan and Shi, Yuanyuan and Xu, Yi and Fu, Luoyi and Zhang, Weinan and Wang, Xinbing and Zhou, Chenghu and Lin, Zhouhan and He, Junxian},
  booktitle = {Proceedings of the 17th ACM International Conference on Web Search and Data Mining (WSDM)},
  pages     = {161--170},
  year      = {2024},
  doi       = {10.1145/3616855.3635772}
}

@article{chen2025jiuzhou,
  title   = {{JiuZhou}: Open Foundation Language Models and Effective Pre-Training Framework for Geoscience},
  author  = {Chen, Zhou and Lin, Ming and Zang, Mingrun and Wang, Zimeng and Li, Juanzi and Bai, Yuqi},
  journal = {International Journal of Digital Earth},
  volume  = {18},
  number  = {1},
  pages   = {2449708},
  year    = {2025},
  doi     = {10.1080/17538947.2025.2449708}
}

@misc{lin2024geogalactica,
  title         = {{GeoGalactica}: A Scientific Large Language Model in Geoscience},
  author        = {Lin, Zhouhan and Deng, Cheng and Zhou, Le and Zhang, Tianhang and Xu, Yi and Xu, Yutong and He, Zhongmou and Shi, Yuanyuan and Dai, Beiya and Song, Yunchong and Zeng, Boyi and Chen, Qiyuan and Miao, Yuxun and Xue, Bo and Wang, Shu and Fu, Luoyi and Zhang, Weinan and He, Junxian and Zhu, Yunqiang and Wang, Xinbing and Zhou, Chenghu},
  year          = {2024},
  eprint        = {2401.00434},
  archivePrefix = {arXiv},
  primaryClass  = {cs.CL},
  note          = {Preprint}
}

@article{ma2026qhgeogpt,
  title   = {A Domain-Specific Language Model for Engineering-Scale Geological Reasoning and Mineral Exploration in the {Qin-Hang} Belt},
  author  = {Ma, Jianhua and Zhou, Yongzhang and Su, Lu and Yang, Huanrong and He, Luhao},
  journal = {Engineering Applications of Artificial Intelligence},
  volume  = {167},
  pages   = {113651},
  year    = {2026},
  doi     = {10.1016/j.engappai.2025.113651}
}

@article{lin2025rockmechanics,
  title   = {Developing a Large Language Model for Oil- and Gas-Related Rock Mechanics: Progress and Challenges},
  author  = {Lin, Botao and Jin, Yan and Cao, Qianwen and Meng, Han and Pang, Huiwen and Wei, Shiming},
  journal = {Natural Gas Industry B},
  volume  = {12},
  pages   = {110--122},
  year    = {2025},
  doi     = {10.1016/j.ngib.2025.03.007}
}

@article{chen2025geofactory,
  title   = {{GeoFactory}: An {LLM} Performance Enhancement Framework for Geoscience Factual and Inferential Tasks},
  author  = {Chen, Zhou and Wang, Xiao and Zhang, Xinan and Lin, Ming and Liao, Yuanhong and Li, Juanzi and Bai, Yuqi},
  journal = {Big Earth Data},
  volume  = {9},
  pages   = {225--248},
  year    = {2025},
  doi     = {10.1080/20964471.2025.2506291}
}

@inproceedings{antoniak2016nlp,
  title     = {Natural Language Processing Techniques on Oil and Gas Drilling Data},
  author    = {Antoniak, M. and Dalgliesh, J. and Verkruyse, M. and Lo, J.},
  booktitle = {SPE Intelligent Energy International Conference and Exhibition},
  year      = {2016},
  doi       = {10.2118/181015-MS}
}

@inproceedings{arumugam2017augmented,
  title     = {Augmented Text Mining for Daily Drilling Reports Using Topic Modeling and Ontology},
  author    = {Arumugam, S. and Rajan, S. and Gupta, S.},
  booktitle = {SPE Western Regional Meeting},
  year      = {2017},
  doi       = {10.2118/185711-MS}
}

@inproceedings{asif2024multilabel,
  title     = {Multi-Label Classification of Daily Drill Reports ({DDR}) Utilizing Large Language Models ({LLMs})},
  author    = {Asif, W. and Al Salt, A. B. and Al Sulaimani, T. and Al Noufli, N.},
  booktitle = {Abu Dhabi International Petroleum Exhibition and Conference (ADIPEC)},
  year      = {2024},
  doi       = {10.2118/221870-MS}
}

@inproceedings{wang2025mudreport,
  title     = {Automatic Daily Drilling Mud Report Processing Using Generative {AI} to Maximize the Operational Efficiency},
  author    = {Wang, J. and Yoon, J. and Marzban, A. and Castanos, R. and Fruge, N. and Holman, I.},
  booktitle = {Offshore Technology Conference (OTC)},
  year      = {2025},
  doi       = {10.4043/35625-MS}
}

@inproceedings{abdelgawad2025finetuned,
  title     = {Fine-Tuned Generative {AI} for Automated Structured Data Extraction and Insight Generation from Legacy Petroleum Well Reports: An {Egyptian} Oilfields Case Study},
  author    = {Abdelgawad, A. E. and El Ghattas, A.},
  booktitle = {Abu Dhabi International Petroleum Exhibition and Conference (ADIPEC)},
  year      = {2025},
  doi       = {10.2118/229443-MS}
}

@inproceedings{eckroth2023petroqa,
  title     = {Answering Natural Language Questions with {OpenAI}'s {GPT} in the Petroleum Industry},
  author    = {Eckroth, J. and Gipson, M. and Boden, J. and Hough, L. and Elliott, J. and Quintana, J.},
  booktitle = {SPE Annual Technical Conference and Exhibition (ATCE)},
  year      = {2023},
  doi       = {10.2118/214888-MS}
}

@inproceedings{pacis2024enhancing,
  title     = {Enhancing Information Retrieval in the Drilling Domain: Zero-Shot Learning with Large Language Models for Question-Answering},
  author    = {Pacis, Felix James and Alyaev, Sergey and Pelfrene, Gilles and Wiktorski, Tomasz},
  booktitle = {IADC/SPE International Drilling Conference and Exhibition},
  year      = {2024},
  doi       = {10.2118/217671-MS}
}

@inproceedings{yi2024applications,
  title     = {Applications of Large Language Models in Well Construction Planning and Real-Time Operation},
  author    = {Yi, M. and Ceglinski, K. and Ashok, P. and Behounek, M. and White, S. and Peroyea, T. and Thetford, T.},
  booktitle = {IADC/SPE International Drilling Conference and Exhibition},
  year      = {2024},
  doi       = {10.2118/217700-MS}
}

@inproceedings{eckroth2025energyllm,
  title     = {Building {EnergyLLM}: A Domain-Specific Large Language Model Trained on {SPE} Content},
  author    = {Eckroth, J. and Boden, J. and Hough, L. and Gatewood, H. and Gipson, S. and Schoen, E. and Gunderson, B.},
  booktitle = {SPE Annual Technical Conference and Exhibition (ATCE)},
  year      = {2025},
  doi       = {10.2118/228097-MS}
}

@article{gpt4_passes,
    author = {Katz, Daniel Martin and Bommarito, Michael James and Gao, Shang and Arredondo, Pablo},
    title = {GPT-4 passes the bar exam},
    journal = {Philosophical Transactions of the Royal Society A: Mathematical, Physical and Engineering Sciences},
    volume = {382},
    number = {2270},
    pages = {20230254},
    year = {2024},
    month = {02},
}

@inproceedings{dettmers2023qlora,
  title     = {{QLoRA}: Efficient Finetuning of Quantized {LLMs}},
  author    = {Dettmers, Tim and Pagnoni, Artidoro and Holtzman, Ari and Zettlemoyer, Luke},
  booktitle = {Advances in Neural Information Processing Systems},
  volume    = {36},
  year      = {2023}
}

@inproceedings{devlin-etal-2019-bert,
    title = "{BERT}: Pre-training of Deep Bidirectional Transformers for Language Understanding",
    author = "Devlin, Jacob  and
      Chang, Ming-Wei  and
      Lee, Kenton  and
      Toutanova, Kristina",
    booktitle = "Proceedings of the 2019 Conference of the North American Chapter of the Association for Computational Linguistics: Human Language Technologies, Volume 1 (Long and Short Papers)",
    year = "2019",
    publisher = "Association for Computational Linguistics",
}

@inproceedings{hoffmann2022training,
  title     = {An Empirical Analysis of Compute-Optimal Large Language Model Training},
  author    = {Hoffmann, Jordan and Borgeaud, Sebastian and Mensch, Arthur and Buchatskaya, Elena and Cai, Trevor and Rutherford, Eliza and de Las Casas, Diego and Hendricks, Lisa Anne and Welbl, Johannes and Clark, Aidan and Hennigan, Tom and Noland, Eric and Millican, Katie and van den Driessche, George and Damoc, Bogdan and Guy, Aurelia and Osindero, Simon and Simonyan, Karen and Elsen, Erich and Vinyals, Oriol and Rae, Jack W. and Sifre, Laurent},
  booktitle = {Advances in Neural Information Processing Systems},
  volume    = {35},
  year      = {2022}
}

@inproceedings{feng2021survey,
  title     = {A Survey of Data Augmentation Approaches for {NLP}},
  author    = {Feng, Steven Y. and Gangal, Varun and Wei, Jason and Chandar, Sarath and Vosoughi, Soroush and Mitamura, Teruko and Hovy, Eduard},
  booktitle = {Findings of the Association for Computational Linguistics: ACL-IJCNLP 2021},
  pages     = {968--988},
  publisher = {Association for Computational Linguistics},
  year      = {2021}
}

\end{document}